# DESARROLLO DE UN DRON LOW-COST PARA TAREAS INDOOR

Mattos Dos Santos, Martin, martinmattos935@gmail.com[1]

Grando Bedin, Ricardo, ricardo.bedin@utec.edu.uy[2]
Kelbouscas, André , andre.dasilva@utec.edu.uy[2]

[1]Escuela Técnica Superior de Rivera, Rivera, Uruguay

[2]Universidad Tecnológica del Uruguay, Rivera, Uruguay

# 4ª FEBITEC — Santana do Livramento e Rivera

**RESUMEN**

Drones comerciales aún no están dimensionados para hacer tareas autónomas indoor, ya que utilizan GPS para su localización en el entorno. Cuando se trata de un espacio con obstáculos físicos (paredes, metal, etc) entre la comunicación del dron y de los satélites que posibilitan la localización precisa de lo mismo, se tiene una gran dificultades de encontrar los satélites o genera interferencia para esta localización. Este problema, puede ocasionar una acción inesperada del dron, pudiendo ocurrir una colisión y un posible accidente,

El trabajo a seguir, presenta el desarrollo de un dron con capacidades de operar en un espacio físico (indoor), sin la necesidad de GPS. En esta propuesta, se desarrolla también un prototipo de un sistema para detección de la distancia (lidar) que el dron está de las paredes, con el objetivo de poder tomar esta información como localización del dron.

*Palabras clave:* *dron indoor, dron lidar, vehículos aéreos no tripulados(VANT)*

**ABSTRACT:**

Commercial drones are not yet dimensioned to perform indoor autonomous tasks, since they use GPS for their location in the environment. When it comes to a space with physical obstacles (walls, metal, etc.) between the communication of the drone and the satellites that allow the precise location of the same, there is great difficulty in finding the satellites or it generates interference for this location. This problem can cause an unexpected action of the drone, a collision and a possible accident can occur,

The work to follow presents the development of a drone capable of operating in a physical space (indoor), without the need for GPS. In this proposal, a prototype of a system for detecting the distance (lidar) that the drone is from the walls is also developed, with the aim of being able to take this information as the location of the drone.

# 1 - INTRODUCCIÓN

Existen características asociadas a vuelos internos con drones, principalmente la problemática de no lograr utilizar GPS en un entorno cerrado. Así como el tamaño, es muy importante en el tema de las estructuras pequeñas, ya que un dron de tamaño grande no entraría en espacios pequeños, también con un tamaño más chico es más difícil ocasionar daño a personas. Otra característica importante de vuelos internos y en espacios colaborativos con personas está la exposición de las hélices, donde (idealmente) tienen que estar protegidas por alguna estructura para protección en el caso de colisionar con alguna persona.

Soluciones de universidades alrededor del mundo para localización en espacios cerrados está en agregar al dron una cámara estéreo, que tiene la capacidad de estimar la posición del dron a partir de las dos cámaras embarcadas en su hardware. Todavía, estas cámaras tienen un costo aproximado desde 1500 dólares, luego el acceso a estos dispositivos está direccionado a investigadores.

Desde ahí, se propone el desarrollo de un dron para servicios indoor con capacidades de funcionar sin gps con utilización de un sistema lidar de bajo costo. Este artículo, presenta la construcción y configuración de un dron para vuelo en ambientes cerrados (*indoor*), con un sistema de detección de obstáculos por medio de sensores de distancia.

# 2 - METODOLOGÍA:

Diagrama de conexión: La señal enviada por el control es recibida por el módulo receptor de radiofrecuencia, los datos recibidos son encaminados a la placa controladora CC3D, esta utiliza cuatro salidas que están conectadas a cuatros ESC, estos son energizados por una batería Lipo de 2200mAh. La funcionalidad de los ESC es energizar los motores con los datos recibidos por la controladora. El diagrama de conexión se puede observar en la Figura 1

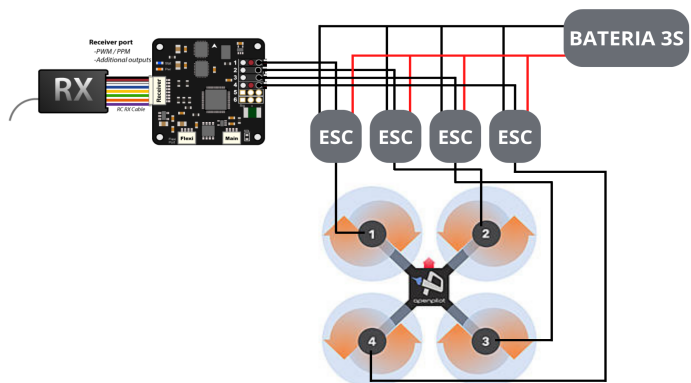

Figura 1. Diagrama de conexión

## 2.1 - HARDWARE NECESARIO

| Item | Descripción | Cantidad | Imagen |
|---|---|---|---|
| Motor Brushless 1806 2400KV | Motor Brushless, es un motor eléctrico y como su propio nombre indica, brushless quiere decir "sin escobillas". | 4 | 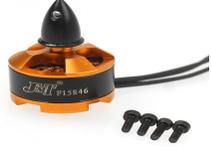 |
| ESC 30A | Los ESC son dispositivos que reciben una señal de la controladora de vuelo y envían la alimentación adecuada a los motores. | 4 | 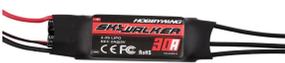 |
| Rádio Transmisor 2.4GHz Turnigy 9X | El radio transmisor es un dispositivo electrónico que con la ayuda de una antena, es capaz de enviar ondas electromagnéticas que pueden contener información. | 1 | 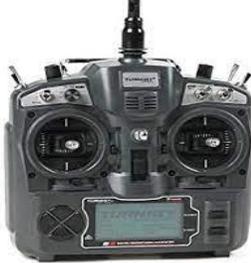 |
| Receptor 2.4GHz FS-R9B | El receptor FS-R9B recibe señales del radio transmisor en la frecuencia de 2.4GHz. | 1 | 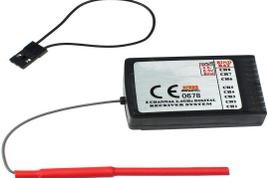 |
| Placa Controladora de Vuelo OpenPilot CC3D | La placa controladora OpenPilot CC3D es un piloto automático (UAV) de código abierto. Es una plataforma altamente capaz para embarcaciones multirrotor, helicópteros y aviones de ala fija. | 1 | 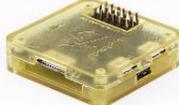 |
| Hélices GEMFAN 3" | Las hélices son el elemento que va a permitir volar al dron, a través de la fuerza que les transmiten los motores. | 4 | 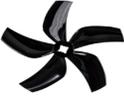 |
| Frame Customizado | El frame customizado es una estructura hecha en impresora 3D Creality Ender 3, para el posicionamiento de los motores y de los controladores, también para la protección de las hélices. | 1 | 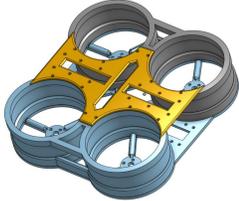 |
| Batería LiPo 3S HPB POWER 2200 mAh | La batería LiPo 3S HPB POWER 2200 mAh es un tipo de batería recargable muy habitual en el mundo de los multirrotores. Nacen como una opción aceptable a la utilización de combustibles para realizar vuelos. | 1 | 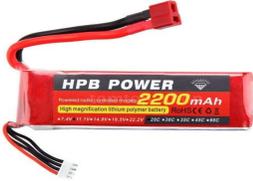 |
| Lidar VL53L1X | Sensor que tiene la capacidad de detectar la distancia a un objeto que esté en el rango de 27 grados . | 12 | 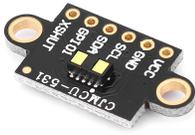 |
| Microcontrolador ESP32 | Procesar la lectura de los sensores Lidar | 1 | 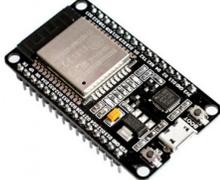 |

## 2.2 - SOFTWARE NECESARIO

Para hacer la gestión de los sensores, y configuración del dron y de la controladora CC3D, se utilizó el software LibrePilot. Este, es un software para proyecto de vehículo aéreo no tripulado libre para modelos de aviones destinado tanto a soportar naves multirrotor como aeronaves de ala fija. En este software se permite configurar los diferentes sensores, actuadores y comunicación que estará embarcado en el dron: Unidad de Medición Inercial (IMU), Controladores Electrónicos de Velocidad (ESC), Motores, Rádio Transmisor y Receptor, además de otras funciones. En la Figura 2 se muestra la interfaz de configuración del software LibrePilot GCS 16.01.

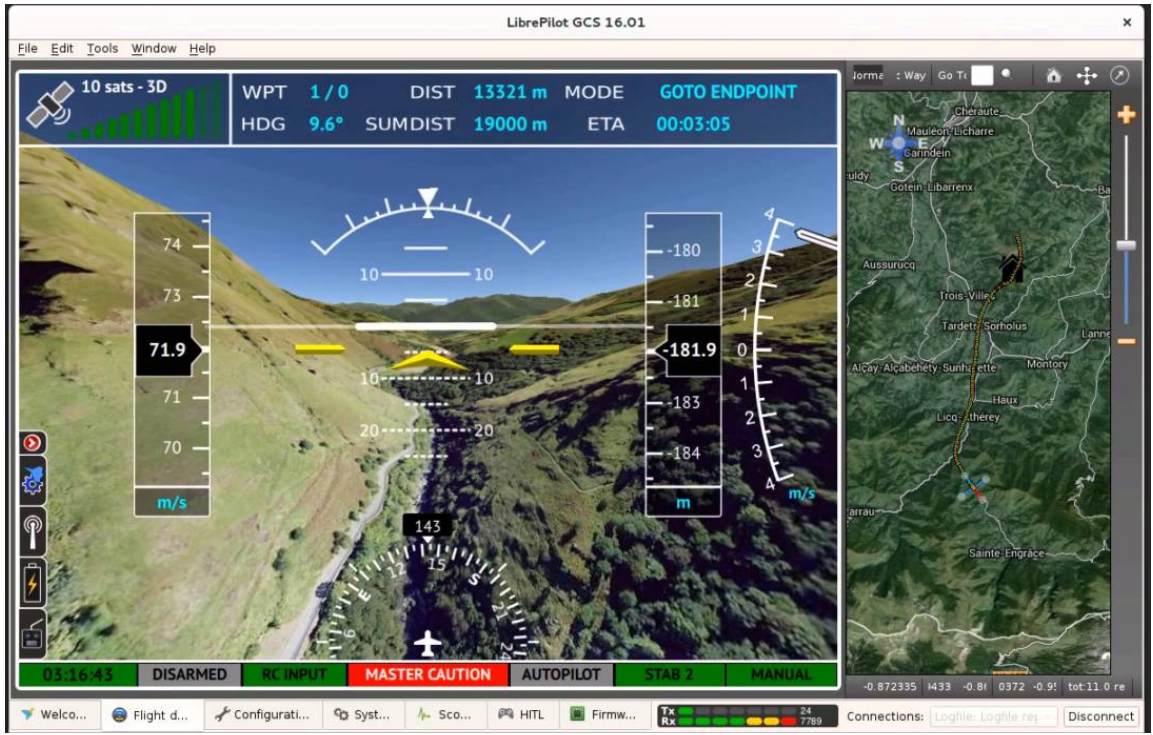

Figura 2. Interfaz de configuración del software LibrePilot GCS 16.01

**3 - RESULTADOS**

Como resultados, se logró desarrollar un dron de dimensión de 250x250mm, como se puede ver en la Figura 2. Fue desarrollado un cable adaptador de una batería para alimentar los 4 ESC. Este cable fue desarrollado con conectores XT60 para no poder conectar de manera equivocada las polaridades de la batería, ya que este conector obliga la conexión en una determinada posición, garantizando la conexión con los polos de la batería y de los ESC de manera correcta.

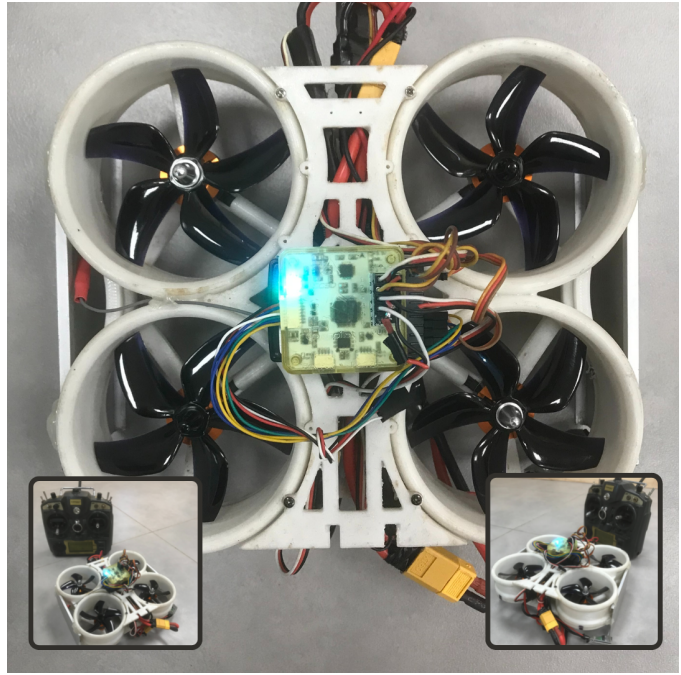

Figura 2. Dron desarrollado por el equipo

Con el software LibrePilot se pudo testear el funcionamiento de los motores a partir de una señal PWM enviada a los ESC. Así como, la Unidad de Medición Inercial(IMU) de la controladora de vuelo OpenPilot CC3D. En el software se hicieron tests y la configuración de la comunicación entre el transmisor (Radio Turnigy 9x 2.4GHz) y receptor (FRSky 2.4GHz) para definición de los canales correspondientes para Yaw, Throttle, Roll y Pitch. Esta configuración fue definida para trabajar con el estándar en aeromodelos llamado Mode 2. Se puede averiguar los comandos y movimientos del aeromodelo con el estándar Mode 2 en la Figura 3.

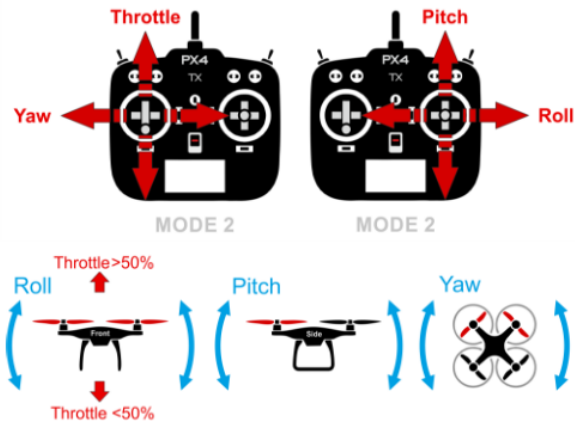

Figura 3. Mode 2 configurado para el modelo

Con eso fue posible el desarrollo de la configuración completa de los sensores, actuadores y comunicación del dron. Para desarrollar la parte de la implementación de un sistema de detección de distancia a objetos y el dron poder localizarse en el ambiente, se utilizó 12 sensores Lidar VL53L1X, con comunicación I2C entre sensores y un microcontrolador ESP32. Se puede visualizar el prototipo en la Figura 4.

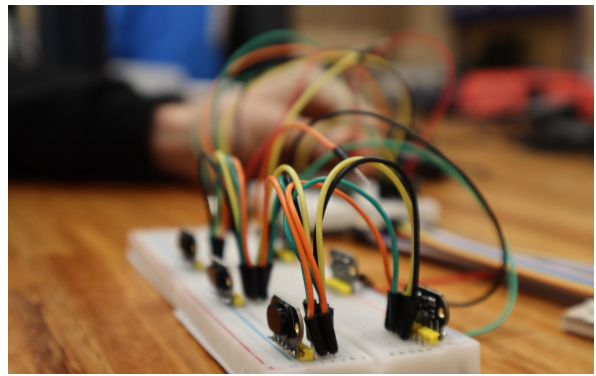

Figura 4. Prototipo de lectura de 6 sensores lidar desde un ESP32 con comunicación I2C

**4 - CONCLUSIÓN**

El presente trabajo tiene como conclusión el uso de protección de las hélices, se realizaron dichas pruebas con las que se pudo ir investigando más a fondo y darle otras mejoras. El dron está preparado para vuelo indoor solo a espera de una red para protección de los integrantes. Los próximos pasos están direccionados a la percepción del dron al ambiente donde está operando. Para eso se está trabajando en la placa para distribuir los sensores Lidar a cada 30 grados del total de una rotación de 360 grados.

El dron tiene un gran potencial para hacer investigación de punta con Inteligencia Artificial y Visión por Computadora. En especial, la comunicación con una torre de procesamiento externo tiene potencial para aumentar la capacidad del dron de navegar con major precisión. Essa comunicación puede ser desarrollada usando placas simples de comunicación como um Raspberry Pi o una ESP 32.

De modo general, tareas indoor son las que tienen major potencial en industria o comercio. La mayoría de las tareas son realizadas en ambientes cerrados y la utilización de vehículos inteligentes puede ser una herramienta para mejorar la capacidad de trabajo de las empresas. Además, la utilización en investigación también tiene potencial, principalmente con investigación relacionada o múltiples drones..

**5 - REFERENCIAS**